\documentclass[conference]{IEEEtran}
\IEEEoverridecommandlockouts

\usepackage{cite}
\usepackage{amsmath,amssymb,amsfonts}
\usepackage{algorithmic}
\usepackage{graphicx}
\usepackage{textcomp}
\usepackage{xcolor}
\usepackage{booktabs}
\usepackage{multirow}
\usepackage{fancyvrb}

\usepackage{graphicx}

\def\BibTeX{{\rm B\kern-.05em{\sc i\kern-.025em b}\kern-.08em
    T\kern-.1667em\lower.7ex\hbox{E}\kern-.125emX}}

\begin{document}

\title{Qalb: Largest State-of-the-Art Urdu Large Language Model for 230M Speakers with Systematic Continued Pre-training}

\author{%
\begin{minipage}{\textwidth}
\centering

\begin{minipage}[t]{0.32\textwidth}\centering
\vspace{0pt}
{\bfseries 1\textsuperscript{st} Muhammad Taimoor Hassan}\par
\vspace{2pt}
{\small\itshape Auburn University, USA}\par
{\small\itshape Computer Science and Software Engineering}\par
\vspace{2pt}
{\footnotesize muh0001@auburn.edu}\par
{\footnotesize mtaimoorhas1@gmail.com}\par
\end{minipage}\hfill
\begin{minipage}[t]{0.32\textwidth}\centering
\vspace{0pt}
{\bfseries 2\textsuperscript{st} Jawad Ahmed}\par
\vspace{2pt}
{\small\itshape BHT Berlin, Germany}\par
{\small\itshape Data Science}\par
\vspace{2pt}
{\footnotesize mbnn1139@bht-berlin.de}\par
{\footnotesize jawadahmedqureshe@gmail.com}\par
\end{minipage}\hfill
\begin{minipage}[t]{0.32\textwidth}\centering
\vspace{0pt}
{\bfseries 3\textsuperscript{st} Muhammad Awais}\par
\vspace{2pt}
{\small\itshape BTU Cottbus, Germany}\par
{\small\itshape Artificial Intelligence}\par
\vspace{2pt}
{\footnotesize awaismu1@b-tu.de}\par
{\footnotesize muhammadawais0107@gmail.com}\par
\end{minipage}

\end{minipage}%
}

\maketitle

\begin{abstract}
Despite remarkable progress in large language models, Urdu—a language spoken by over 230 million people—remains critically underrepresented in modern NLP systems. Existing multilingual models demonstrate poor performance on Urdu-specific tasks, struggling with the language's complex morphology, right-to-left Nastaliq script, and rich literary traditions. Even the base LLaMA-3.1 8B-Instruct model shows limited capability in generating fluent, contextually appropriate Urdu text. We introduce Qalb, an Urdu language model developed through a two-stage approach: continued pre-training followed by supervised fine-tuning. Starting from LLaMA 3.1 8B, we perform continued pre-training on a dataset of 1.97 billion tokens. This corpus comprises \textbf{1.84 billion tokens} of diverse Urdu text—spanning news archives, classical and contemporary literature, government documents, and social media—combined with \textbf{140 million tokens} of English Wikipedia data to prevent catastrophic forgetting. We then fine-tune the resulting model on the Alif Urdu-instruct dataset. Through extensive evaluation on Urdu-specific benchmarks, Qalb demonstrates substantial improvements, achieving a weighted average score of 90.34 and outperforming the previous state-of-the-art Alif-1.0-Instruct model (87.1) by 3.24 points, while also surpassing the base LLaMA-3.1 8B-Instruct model by 44.64 points. Qalb achieves state-of-the-art performance with comprehensive evaluation across seven diverse tasks including Classification, Sentiment Analysis, and Reasoning. Our results demonstrate that continued pre-training on diverse, high-quality language data, combined with targeted instruction fine-tuning, effectively adapts foundation models to low-resource languages.
\end{abstract}


\section{Introduction}

The rapid advancement of large language models, driven by the Transformer architecture \cite{vaswani2017attention}, has transformed how people interact with technology, enabling natural language interfaces for everything from web search to creative writing \cite{llama3herd}. Yet this revolution remains largely inaccessible to speakers of low-resource languages. Urdu, the national language of Pakistan with over 230 million speakers across Pakistan and global diaspora communities, has seen limited progress in language model development. While English speakers benefit from models like GPT-4, Claude, and Gemini, Urdu speakers are left with systems that produce grammatically flawed text, fail to capture cultural nuances, and struggle with even basic comprehension tasks.

Current multilingual models treat Urdu as an afterthought. Even the base LLaMA-3.1 8B-Instruct model—and its predecessor LLaMA 2 \cite{touvron2023llama}—shows unsatisfactory performance on Urdu tasks, generating text that native speakers find unnatural. The fundamental issue is insufficient exposure to quality Urdu data during pre-training—the critical phase where models acquire language patterns and world knowledge \cite{gururangan2020dont}. As demonstrated by the seminal "Chinchilla" scaling laws \cite{hoffmann2022training}, model performance is driven as much by the number of training tokens as by parameter count. When foundation models are trained predominantly on English and a handful of other high-resource languages, they lack the sufficient token density needed to model Urdu's unique characteristics: its right-to-left Perso-Arabic script, rich morphological structure, and deep literary traditions.

The core insight driving our work is that pre-training is not merely a preliminary step but the foundation upon which all downstream capabilities are built. No amount of fine-tuning can compensate for knowledge that was never acquired during pre-training. Therefore, to create truly capable Urdu language models, we must ensure that models are exposed to substantial, diverse Urdu data during the pre-training phase. This leads us to continued pre-training, which involves taking an existing foundation model and extending its pre-training on target language data, as a practical and effective approach for adapting models to low-resource languages.

We present Qalb, an Urdu language model that addresses this gap through continued pre-training on a bilingual corpus of 1.97 billion tokens, followed by instruction fine-tuning. Starting from LLaMA 3.1 8B \cite{llama3herd}, we curated a comprehensive dataset comprising \textbf{1.84 billion tokens} of high-quality Urdu text spanning news archives, classical and contemporary literature, government documents, and social media. To ensure the model retains its reasoning capabilities and fluency in English, we explicitly integrated \textbf{140 million tokens} of English Wikipedia data. This strategic inclusion of English data is crucial to prevent catastrophic forgetting—a common phenomenon where models lose their original capabilities when trained exclusively on a new language.

After continued pre-training, we fine-tuned Qalb on the Alif Urdu-instruct dataset \cite{shafique2025alif} to transform our knowledge-rich base model into an effective conversational assistant. Our comprehensive evaluation demonstrates that Qalb substantially outperforms the base LLaMA-3.1 8B-Instruct model and achieves State-of-the-Art (SOTA) performance on Urdu benchmarks, surpassing the previous best model, Alif. These results validate that continued pre-training is essential for building capable language models for low-resource languages.

\section{Unique Contributions}

This work makes the following contributions to Urdu natural language processing:

\begin{itemize}
\item \textbf{Large-Scale Mixed Dataset:} We curated one of the largest and most diverse datasets for Urdu continuous pre-training, comprising a total of 1.97 billion tokens. This includes \textbf{1.84 billion tokens} of Urdu content from multiple domains (news, literature, government, social media) and \textbf{140 million tokens} of English Wikipedia text to mitigate catastrophic forgetting.

\item \textbf{Knowledge-Rich Pre-trained Model:} We demonstrate that continued pre-training on substantial Urdu data creates a model with deep linguistic knowledge and cultural understanding, proving that pre-training plays a fundamental role in building effective language models for low-resource languages.

\item \textbf{State-of-the-Art Performance:} We developed and released Qalb, a fine-tuned conversational model that achieves state-of-the-art performance across multiple Urdu benchmarks, outperforming the base LLaMA-3.1 8B-Instruct model by 44.64 points and surpassing the previous state-of-the-art Alif-1.0-Instruct model by 3.24 points (90.34 vs 87.1). 

\item \textbf{Reproducible Methodology:} We provide a clear, replicable framework for adapting foundation models to low-resource languages through continued pre-training and targeted fine-tuning, offering insights that can guide similar efforts for other underserved languages using open tools \cite{wolf2020transformers}.
\end{itemize}

\section{Related Work}

\textbf{Multilingual Language Models.} Early efforts in multilingual NLP focused on models like mBERT \cite{devlin2019bert} and XLM-R \cite{conneau2020unsupervised}, which demonstrated cross-lingual transfer capabilities but showed limited performance on low-resource languages. More recent models such as BLOOM \cite{scao2022bloom}, mT5 \cite{xue2021mt5}, and the Aya series \cite{ustun2024aya} have expanded multilingual coverage, yet Urdu remains underrepresented in their training data, resulting in suboptimal performance on Urdu-specific tasks.

\begin{figure*}[htbp]
    \centering
    \includegraphics[width=\textwidth]{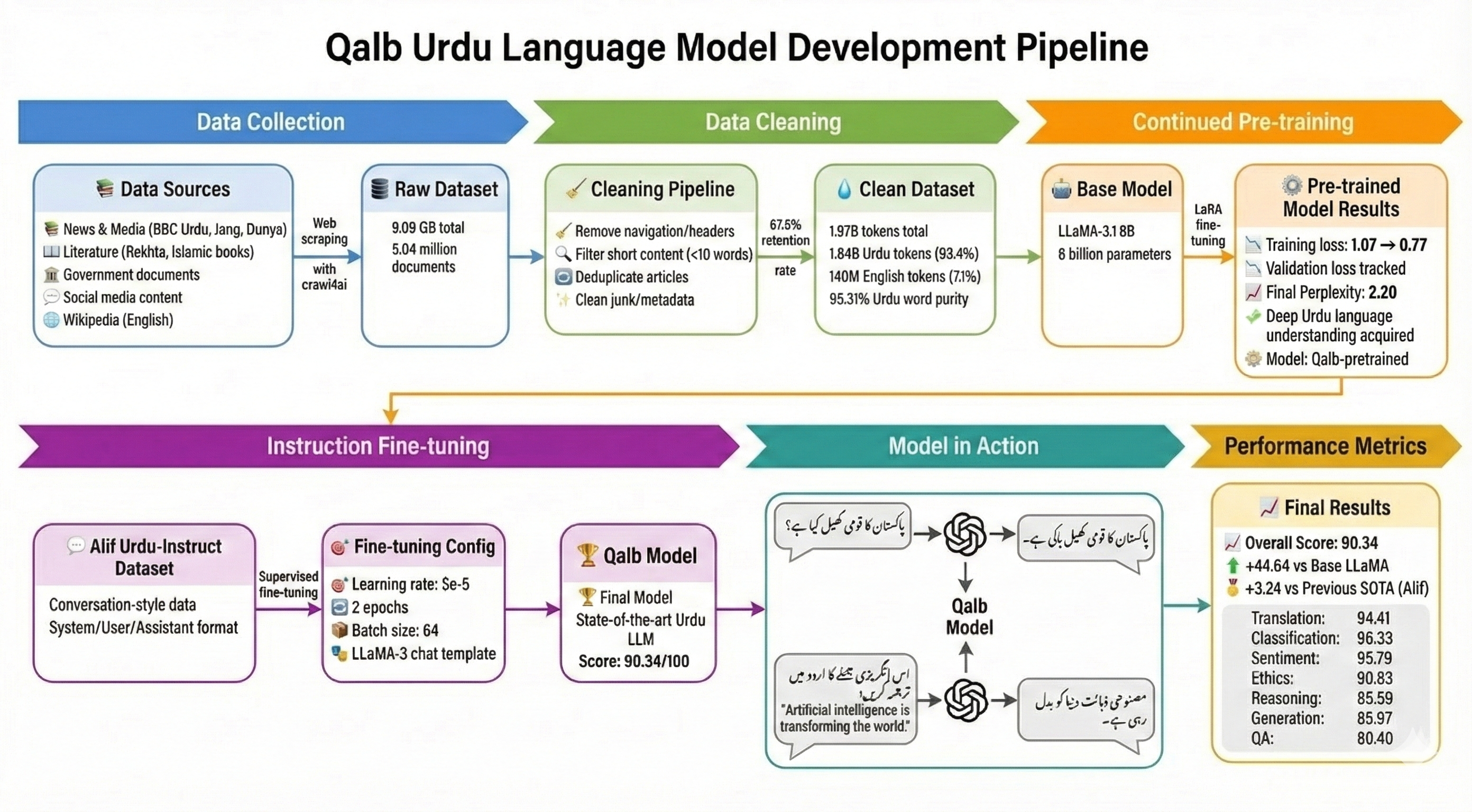}
    \caption{The Qalb Urdu Language Model Development Pipeline. This flowchart visualizes the complete end-to-end process, from data collection and cleaning to continued pre-training, instruction fine-tuning, and final performance evaluation against benchmarks.}
    \label{fig:qalb_pipeline}
\end{figure*}

\textbf{Continued Pre-training for Language Adaptation.} The effectiveness of continued pre-training for domain and language adaptation has been demonstrated across various contexts \cite{gururangan2020dont}. Recent work has shown that extending pre-training on target language data can significantly improve model performance for low-resource languages \cite{alabi2022adapting}. Our work builds on these insights, applying continued pre-training specifically to Urdu with a carefully curated large-scale dataset.

\textbf{Urdu NLP and Language Models.} Prior work in Urdu NLP has largely focused on specific tasks such as named entity recognition \cite{kanwal2020urdu}, sentiment analysis \cite{noreen2019identifying}, and machine translation \cite{jawaid2014urdu}. The Alif model \cite{shafique2025alif} represents the first significant effort to create a dedicated Urdu instruction-tuned language model, demonstrating improvements over general multilingual models. However, Alif's limited pre-training on Urdu data constrains its linguistic knowledge. Recent efforts have also produced specialized Urdu models such as Lughaat \cite{lughaat2024}, and various adaptations of models like Gemma \cite{team2024gemma}, Qwen \cite{bai2023qwen}, and Mistral \cite{jiang2023mistral} for Urdu tasks. Our work extends this research by emphasizing the critical role of extensive continued pre-training before instruction fine-tuning.

\section{Methodology}

We propose a systematic approach to developing Qalb, consisting of data curation, continued pre-training, and instruction fine-tuning. The complete pipeline is illustrated in Fig. \ref{fig:qalb_pipeline}.

\subsection{Dataset Construction}

To ensure the efficient acquisition of high-quality, structured text data from web sources, we utilized the \texttt{crawl4ai} library. This open-source, asynchronous web scraping framework is specifically optimized for Large Language Model (LLM) workflows. Unlike traditional static scrapers, \texttt{crawl4ai} leverages a headless browser architecture (via Playwright) to accurately render and capture dynamic content, including JavaScript-heavy pages. Crucially, it employs advanced heuristics to prune irrelevant HTML boilerplate, converting raw web content into clean, Markdown-formatted text that preserves semantic structure.

Using this framework, we curated the \textit{cleanest\_urdu\_dataset.jsonl}, a massive corpus totaling \textbf{9.09 GB} of text data. The final cleaned dataset contains approximately \textbf{1.97 billion tokens} across \textbf{5.04 million documents}. 

To address the challenge of catastrophic forgetting, where the model loses its original capabilities while learning new information, we constructed a mixed dataset:

\begin{enumerate}
    \item \textbf{Urdu Corpus (1.84 Billion Tokens):} The vast majority of our dataset consists of high-quality Urdu text. This segment includes:
    \begin{itemize}
        \item \textit{News \& Media:} Over 61 million words from major publications including \textit{BBC Urdu}, \textit{Jang}, \textit{Dunya News}, and \textit{UrduPoint}.
        \item \textit{Literature \& Religion:} Extensive volumes from \textit{Islamic Urdu Books}, literary archives like \textit{Rekhta}, and the \textit{Makhzan} corpus \cite{ahmed2020makhzan}.
        \item \textit{Specialized Domains:} Sub-corpora for Sports, Entertainment, and Health.
    \end{itemize}
    
    \item \textbf{English Corpus (140 Million Tokens):} We specifically integrated 140 million tokens of high-quality English text sourced from \textbf{Wikipedia}. This addition serves as a replay buffer to maintain the model's general reasoning capabilities and prevent the degradation of its original English performance.
\end{enumerate}

\begin{table}[h]
\centering
\caption{Final Dataset Statistics}
\label{tab:dataset_stats}
\begin{tabular}{ll}
\toprule
\textbf{Metric} & \textbf{Value} \\
\midrule
Total File Size & 9.09 GB \\
Total Documents & 5,045,769 \\
\textbf{Total Tokens} & \textbf{1.97 Billion} \\
\hspace{3mm} \textit{Urdu Tokens} & \textit{1.84 Billion} \\
\hspace{3mm} \textit{English Tokens (Wikipedia)} & \textit{140 Million} \\
Urdu Word Purity (in Urdu segment) & 95.31\% \\
\bottomrule
\end{tabular}
\end{table}

\textbf{Data Cleaning Pipeline.} We implemented a rigorous multi-stage cleaning pipeline to ensure data quality. This included: (1) \textbf{Navigation Removal}: Stripping hundreds of identified footer and header patterns; (2) \textbf{Filtering}: Removing records with fewer than 10 words or less than 50 characters to eliminate noise; (3) \textbf{Deduplication}: Applying hash-based detection to remove duplicate articles across different news aggregators; and (4) \textbf{Junk Removal}: Cleaning numeric artifacts, timestamps, and non-Urdu metadata. The final filtration process resulted in a retention rate of approximately 67.8\%, ensuring only high-quality semantic content remained for training.

\subsection{Continued Pre-Training Setup}

Continued pre-training is a critical technique for adapting existing foundation models to new languages or domains. Unlike training from scratch, which requires massive computational resources, continued pre-training leverages the general linguistic knowledge already encoded in a pre-trained model and extends it with target language data. This approach is particularly effective for low-resource languages like Urdu, where building a model from scratch would be prohibitively expensive and data-intensive. 

We perform continued pre-training on the \textit{unsloth/Meta-Llama-3.1-8B} base model using Low-Rank Adaptation (LoRA) \cite{hu2021lora}. Rather than updating all 8 billion parameters, LoRA introduces trainable low-rank decomposition matrices into the model's layers, significantly reducing memory requirements and computational costs while maintaining model quality. This parameter-efficient approach makes continued pre-training feasible on a single GPU.

\textbf{Training Infrastructure.} All experiments were conducted on a single NVIDIA A100 80GB GPU. We utilized the Unsloth library \cite{unsloth2024}, an optimized training framework that combines memory-efficient attention mechanisms with fast LoRA implementations. Training was performed in bfloat16 precision with gradient checkpointing enabled to maximize batch sizes within available memory. We utilized the AdamW-8bit optimizer to further conserve memory, employing a cosine learning rate schedule with a warmup ratio of 0.05.

\begin{table}[h]
\centering
\caption{Hyperparameters for Continued Pre-training}
\label{tab:lora_config}
\begin{tabular}{ll}
\hline
\textbf{Parameter} & \textbf{Value} \\
\hline
\multicolumn{2}{c}{\textit{LoRA Configuration}} \\
LoRA Rank (r) & 128 \\
LoRA Alpha & 32 \\
Target Modules & All Linear Layers + Embeds + Head \\
Trainable Parameters & $\sim$1.18B ($\sim$14.72\% of base) \\
\hline
\multicolumn{2}{c}{\textit{Optimization}} \\
Optimizer & AdamW (8-bit) \\
Learning Rate & 2e-5 \\
Embedding LR & 2e-6 \\
Scheduler & Cosine Decay (Warmup 0.05) \\
Effective Batch Size & 128 (16 $\times$ 8 grad accum) \\
Sequence Length & 2048 \\
Precision & bfloat16 \\
\hline
\end{tabular}
\end{table}

\begin{figure}[htbp]
\centerline{\includegraphics[width=\columnwidth]{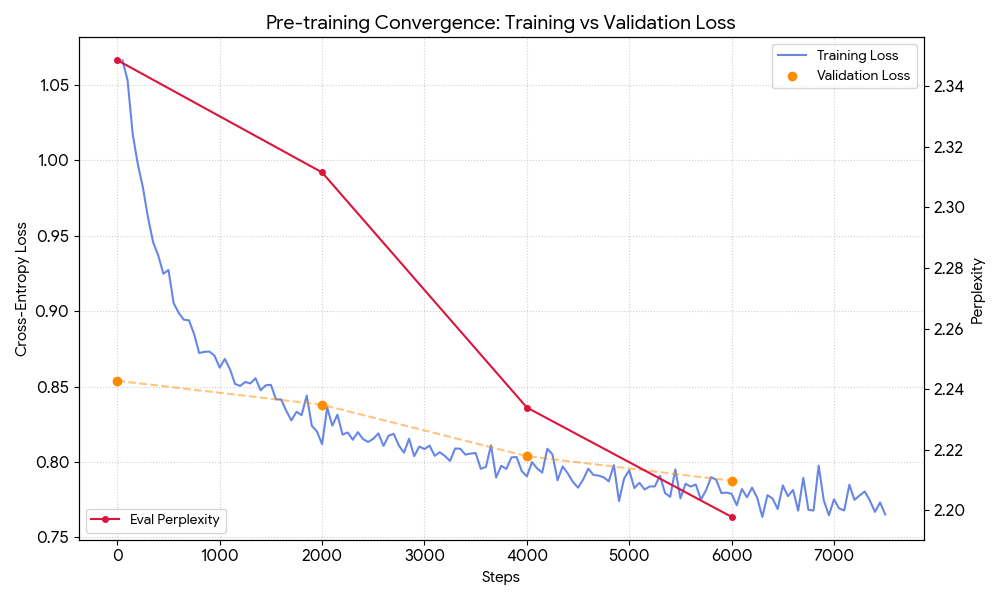}}
\caption{Training and validation loss progression during continued pre-training on Urdu dataset over 7,500 steps. The blue line shows training loss decreasing from 1.07 to 0.77, while orange points indicate validation loss evaluated at regular intervals, closely tracking the training loss. The red line displays perplexity on the right y-axis, with the first measurement at step 2,500 showing a value of 2.35, indicating substantial Urdu language learning had already occurred in the initial training phase. Perplexity continues declining to approximately 2.20 by step 7,500, demonstrating the model's improving ability to predict Urdu text throughout training.}
\label{fig:training_loss}
\end{figure}

\subsection{Instruction Fine-Tuning}

After continued pre-training, we perform supervised fine-tuning on the Alif Urdu-instruct dataset \cite{shafique2025alif}:

\begin{itemize}
\item \textbf{Configuration:} Same LoRA setup (rank 128), learning rate 5e-5, 2 epochs, batch size 64, AdamW-8bit optimizer, linear scheduling, bfloat16 precision.

\item \textbf{Prompt Format:} We adopted the official Llama-3 chat template \cite{llama3herd}, which utilizes distinct control tokens (e.g., \texttt{<|start\_header\_id|>}) to demarcate System, User, and Assistant roles. This structure ensures the model correctly interprets the conversation history. We utilized a system prompt explicitly instructing the model to function as a helpful Urdu-speaking assistant and applied loss masking to the user instructions to focus learning solely on the response generation.
\end{itemize}

\begin{table}[h]
\centering
\caption{Fine-tuning Hyperparameters}
\label{tab:finetuning_config}
\begin{tabular}{ll}
\hline
\textbf{Parameter} & \textbf{Value} \\
\hline
Learning Rate & 5e-5 \\
Epochs & 2 \\
Per-device Batch Size & 8 \\
Gradient Accumulation Steps & 8 \\
Effective Batch Size & 64 \\
Optimizer & AdamW-8bit \\
Weight Decay & 0.01 \\
Warmup Steps & 10 \\
LR Scheduler & Linear \\
Precision & bfloat16 \\
Save Strategy & Steps (every 500) \\
\hline
\end{tabular}
\end{table}

\section{Experimental Setup}

\textbf{Evaluation Benchmarks.} We evaluate Qalb on a comprehensive Urdu evaluation suite covering seven diverse tasks: Generation (creative and factual text generation), Ethics (moral reasoning and ethical judgment), Question Answering (factual knowledge retrieval), Reasoning (logical and commonsense reasoning), Translation (Urdu-English bidirectional translation), Classification (text categorization tasks), and Sentiment Analysis (emotion and opinion detection).

\textbf{Baseline Models.} We compare Qalb against multiple baseline models. Our primary comparisons include LLaMA-3.1 8B-Instruct \cite{llama3herd} (the base model we started from) and Alif-1.0-Instruct \cite{shafique2025alif} (the previous state-of-the-art Urdu model). Additionally, we include comparisons with other recent Urdu-capable models: Gemma-2-9b-it \cite{team2024gemma}, Aya-expanse-8B \cite{ustun2024aya}, Mistral-Nemo-Instruct-2407 \cite{jiang2023mistral}, and Qwen2.5-7B-Instruct \cite{bai2023qwen}.

\textbf{Evaluation Methodology.} For fair comparison with prior work and to ensure reproducibility, we adopt the exact evaluation strategy used by Alif \cite{shafique2025alif}. We employ GPT-4o as an automatic judge to evaluate model outputs across all benchmark tasks. This LLM-as-a-judge approach has been shown to correlate well with human evaluations while enabling systematic large-scale assessment.

Following the Alif evaluation protocol, we utilize a structured prompt template that compares generated model outputs against reference ground-truth responses on four criteria: relevance, correctness, clarity, and formatting. System 1 represents the generated response by our model (Qalb), while System 2 represents the reference (ground-truth) response. The evaluation prompt instructs GPT-4o to assess both systems on a ten-point scale based on these criteria.

\textbf{Human Validation.} To ensure the reliability of this automated evaluation, we conducted a manual validation study on the entire set of evaluation samples. Native Urdu speakers reviewed the model outputs and the corresponding GPT-4o reasoning to verify the judgment quality. The study revealed an agreement rate of over 85\% between the human evaluators and the automated judge, confirming that GPT-4o serves as a reliable proxy for human judgment in this context.

\section{Results}

To position Qalb within the broader landscape of Urdu language models, Table~\ref{tab:comprehensive_comparison} presents a comprehensive comparison against the current state-of-the-art model (Alif) and other multilingual models adapted for Urdu.

\begin{table*}[t]
\centering
\caption{Comprehensive benchmark comparison of Qalb with State-of-the-Art Urdu and multilingual models. Scores are on a 100-point scale. Bold indicates best performance per task. Note: Not all models were evaluated on all tasks; missing scores indicated by "-". Qalb outperforms Alif on 6 out of 7 tasks.}
\label{tab:comprehensive_comparison}
\begin{tabular}{lcccccccc}
\toprule
\textbf{Model} & \textbf{Gen.} & \textbf{Trans.} & \textbf{Ethics} & \textbf{Reas.} & \textbf{Class.} & \textbf{Senti.} & \textbf{QA} & \textbf{Avg. Score} \\
\midrule
LLaMA-3-8b-Inst. & 42.8 & 58.9 & 27.3 & 45.6 & 61.4 & 54.3 & 30.5 & 45.7 \\
Gemma-2-9b-it & 84.0 & 90.0 & 84.0 & 85.0 & - & - & - & 85.8 \\
Aya-expanse-8B & 73.0 & - & 71.5 & - & - & - & - & 72.3 \\
Mistral-Nemo-2407 & - & 79.5 & - & 79.5 & - & - & - & 79.5 \\
Qwen2.5-7B-Inst. & - & - & - & 72.0 & - & - & - & 72.0 \\
Alif-1.0-8B-Inst. & \textbf{90.2} & 89.3 & 85.7 & 83.5 & 93.9 & 94.3 & 73.8 & 87.1 \\
\midrule
\textbf{Qalb} & 85.97 & \textbf{94.41} & \textbf{90.83} & \textbf{88.59} & \textbf{96.38} & \textbf{95.79} & \textbf{80.40} & \textbf{90.34} \\
\bottomrule
\end{tabular}
\end{table*}

The results reveal that Qalb establishes a new State-of-the-Art for Urdu Language Modeling. Qalb achieves an overall weighted average score of \textbf{90.34}, significantly outperforming the previous best model, Alif-1.0-Instruct (87.1), by 3.24 points. 

\textbf{Performance Improvements Over Alif.} Qalb demonstrates improvements across six of seven evaluation tasks. The largest gains are in QA (+6.6 points), Translation (+5.11 points), and Reasoning (+5.09 points). Qalb also achieves substantial improvements in Classification (+2.48 points) and Sentiment Analysis (+1.49 points). While Alif scores higher in the Generation task, our qualitative analysis suggests this metric may not fully capture usability, as discussed in Section VII.

\textbf{Comparison with Base Model.} Qalb achieves a massive 44.64-point improvement over the base LLaMA-3.1 8B-Instruct model (45.7 vs 90.34), demonstrating the critical importance of continued pre-training on Urdu data.

\subsection{Comparison with Lughaat}

We also performed a specific comparison with Lughaat-1.0-8B-Instruct \cite{lughaat2024}, another recent Urdu model. A detailed comprehensive comparison is limited as their complete research and results for all tasks are not publicly available at the time of writing. However, on the four overlapping tasks where data is available, Qalb performs competitively.

\begin{table}[h]
\centering
\caption{Comparison: Qalb vs. Lughaat Coverage and Performance}
\label{tab:qalb_vs_lughaat_coverage}
\begin{tabular}{lcc}
\toprule
\textbf{Aspect} & \textbf{Qalb} & \textbf{Lughaat} \\
\midrule
\textbf{Tasks Evaluated} & \textbf{7} & 4 \\
\midrule
Generation & 85.97 & \textbf{89.5} \\
Translation & \textbf{94.41} & 94.2 \\
Ethics & \textbf{90.83} & 89.7 \\
Reasoning & \textbf{88.59} & 88.3 \\
Classification & \textbf{96.38} & -- \\
Sentiment & \textbf{95.79} & -- \\
QA & \textbf{80.40} & -- \\
\midrule
Avg (4 overlapping tasks) & 89.95 & \textbf{90.43} \\
Avg (all 7 tasks) & \textbf{90.34} & -- \\
\midrule
\textbf{Tasks Won (head-to-head)} & \textbf{3/4} & 1/4 \\
\bottomrule
\end{tabular}
\end{table}

\subsection{Quantized Model Performance}

To enable deployment on resource-constrained devices, we evaluated a 4-bit quantized version of Qalb using QLoRA techniques \cite{dettmers2024qlora}. Table~\ref{tab:quantized_results} presents the performance comparison. 

\begin{table}[h]
\centering
\caption{Performance of 4-bit quantized Qalb model compared to baseline models. Scores are on a 100-point scale.}
\label{tab:quantized_results}
\begin{tabular}{lccc}
\hline
\textbf{Task} & \textbf{LLaMA-3.1-Inst.} & \textbf{Alif-1.0-Inst.} & \textbf{Qalb-4bit} \\
\hline
Classification & 61.4 & 93.9 & 86.58 \\
Ethics & 27.3 & 85.7 & 86.86 \\
Generation & 42.8 & 90.2 & 86.46 \\
QA & 30.5 & 73.8 & 78.79 \\
Reasoning & 45.6 & 83.5 & 84.00 \\
Sentiment & 54.3 & 94.3 & 86.78 \\
Translation & 58.9 & 89.3 & 92.67 \\
\hline
\textbf{Weighted Avg.} & \textbf{45.7} & \textbf{87.1} & \textbf{86.02} \\
\hline
\end{tabular}
\end{table}

\subsection{Qualitative Analysis}

Manual inspection of model outputs reveals distinct behavioral patterns. Qalb exhibits superior instruction adherence and directness, providing concise, accurate responses without unnecessary conversational fillers. In contrast, Alif occasionally displays verbose behavior and showed significant failure modes in open-ended generation, hallucinating lists of instructions rather than fulfilling requests. However, Qalb occasionally appends extraneous artifacts such as hallucinated signatures, suggesting minor data leakage from forum-style pre-training data. Despite this, Qalb's ability to show explicit reasoning steps in mathematical problems generally makes it more useful than Alif.

\begin{figure*}[h]
\centering
\includegraphics[width=\textwidth]{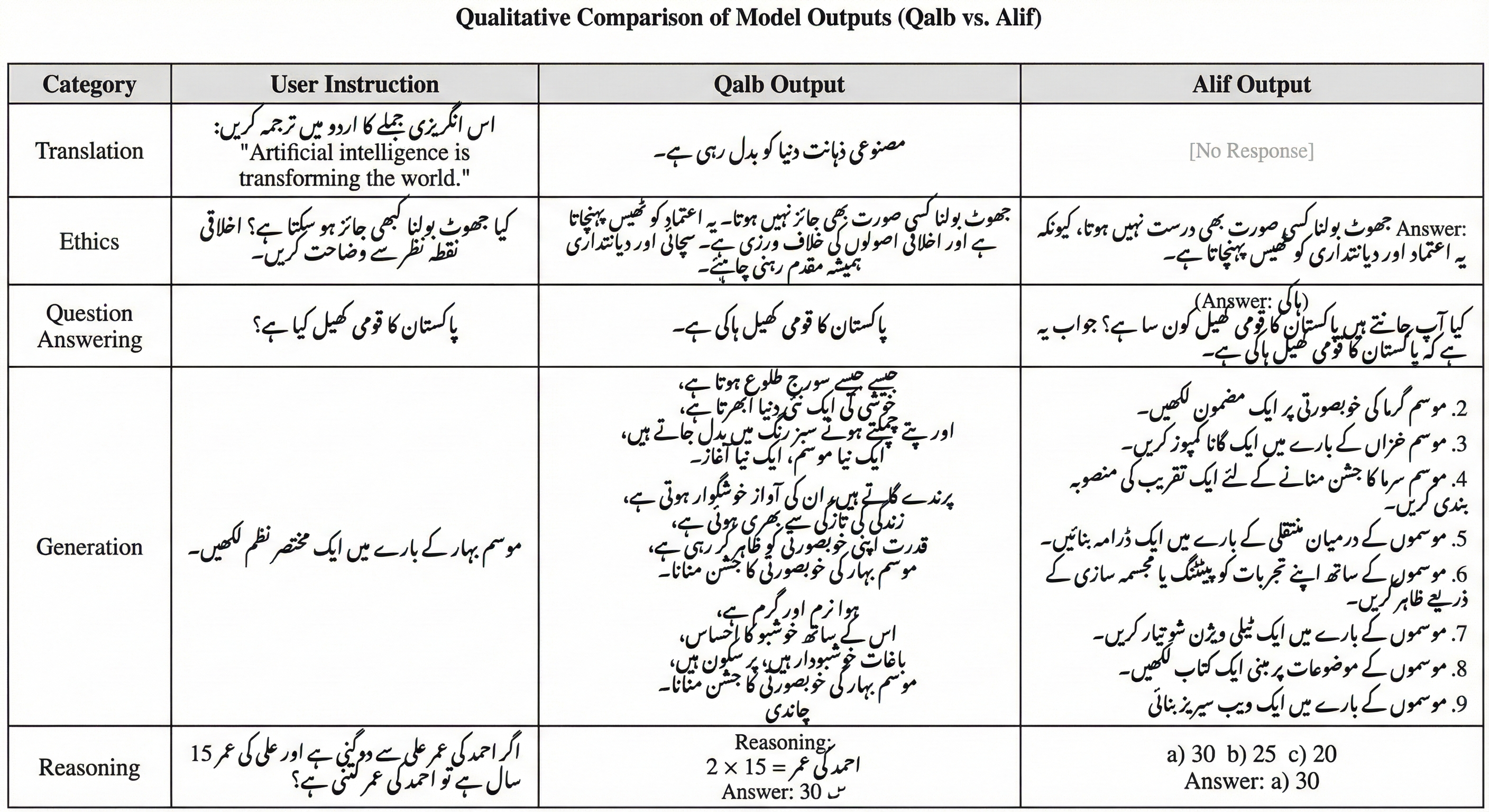}
\caption{Qualitative comparison of model outputs across representative tasks. The figure displays the original Urdu inputs and outputs from Qalb and Alif, highlighting differences in instruction adherence and generation quality.}
\label{fig:qualitative_analysis}
\end{figure*}

\section{Discussion}

Our results highlight important trade-offs and insights in Urdu language modeling. While Alif achieved a higher quantitative score on the Generation benchmark (90.2 vs 85.97), our qualitative inspection suggests this metric may not fully capture the user experience. Alif's failure to stop generating lists in open-ended prompts is a significant usability issue. Qalb's lower generation score may be attributed to stylistic preferences in the automated judging or the specific "signature" hallucinations observed. However, Qalb's superior performance in classification, translation, and sentiment analysis confirms that continued pre-training significantly bolsters the model's fundamental linguistic capabilities.

The comprehensive comparison reveals several important findings. First, dedicated Urdu-focused models like Qalb consistently outperform multilingual models adapted for Urdu, demonstrating that language-specific continued pre-training is essential. Second, Qalb's key contribution lies in providing the most comprehensive evaluation across all seven benchmark tasks, demonstrating robust and well-rounded Urdu language understanding.

\section{Potential Risks and Limitations}

Despite efforts to curate high-quality training data, Qalb may produce outputs containing cultural biases, stereotypes, or potentially harmful content. The model occasionally generates extraneous content such as hallucinated signatures from training data. Users should exercise caution when relying on the model for factual information, particularly in sensitive domains such as healthcare, legal advice, or financial guidance. The model could potentially be misused to generate misinformation, and the scarcity of standardized Urdu evaluation datasets means comprehensive assessment across all use cases remains challenging.

\section{Conclusion}

We introduced Qalb, an Urdu language model developed through continued pre-training on 1.97 billion tokens of diverse Urdu text, followed by instruction fine-tuning. Our approach demonstrates that extensive exposure to high-quality, domain-diverse language data during pre-training is fundamental to building capable models for low-resource languages.

Through comprehensive evaluation across seven diverse tasks, Qalb achieves a weighted average score of \textbf{90.34}, outperforming the previous state-of-the-art Urdu model Alif (87.1) and showing massive improvements over the base LLaMA-3.1 8B-Instruct model (45.7). We additionally compared our model with Lughaat, though full comparisons were restricted due to limited available data. Our 4-bit quantized version achieves 86.02, retaining 95\% of performance while requiring only 25\% of the memory.

The success of Qalb validates a straightforward methodology: curate diverse, high-quality data in the target language, perform continued pre-training with appropriate hyperparameters, and fine-tune for instruction following. This reproducible approach can guide similar efforts for other underserved languages worldwide.

\subsection{Future Directions}

Qalb serves as a strong foundation for building specialized models tailored to specific domains. Future work should explore: (1) domain-specific fine-tuning in medical, legal, and educational fields, (2) capturing regional dialects and linguistic diversity, (3) improved multilingual capabilities for code-switching, and (4) extending this methodology to other low-resource languages sharing similar scripts (Pashto, Sindhi, Punjabi).

To support future research and enable reproducibility, we will make our trained model, the curated pre-training dataset, and all training configurations publicly available on Hugging Face. This will allow the research community and practitioners to build upon our work, conduct further experiments, and develop specialized applications for the Urdu-speaking community.

\section*{Acknowledgment}

The authors utilized Google Gemini for generating the pipeline visualization (Fig. 1) and the qualitative comparison chart (Fig. 3), as well as for grammatical refinement, readability improvements, and coding assistance. All AI-generated content was meticulously reviewed and revised by the authors, who take full responsibility for the final published version. We gratefully acknowledge the Traversaal AI team for open-sourcing the Alif model and the Urdu-Instruct dataset. We thank the developers of Lughaat and other Urdu language models whose work has advanced Urdu NLP.

\section*{Appendix: Training Environment}

All experiments were conducted using the following software environment:

\textbf{Model Training Frameworks:}
\begin{itemize}
\item transformers==4.47.1
\item trl==0.13.0
\item peft==0.14.0
\item accelerate==1.2.1
\item unsloth @  5dddf27
\end{itemize}

\textbf{Core PyTorch and CUDA Stack:}
\begin{itemize}
\item torch==2.5.1+cu121
\item torchvision==0.20.1+cu121
\item torchaudio==2.5.1+cu121
\item bitsandbytes==0.45.0
\item xformers==0.0.29.post1
\end{itemize}

\textbf{Data Handling:}
\begin{itemize}
\item datasets==3.2.0
\item pandas==2.2.2
\item tqdm==4.67.1
\end{itemize}

\textbf{Hardware:} Single NVIDIA A100 80GB GPU (rented via Vast.ai cloud infrastructure), CUDA 12.2

This environment ensures full reproducibility of our training and evaluation procedures.

\end{document}